\documentclass[10pt, journal,onecolumn]{IEEEtran}

\usepackage{amsmath}
\input{amssym}
\usepackage{graphicx}
\usepackage{caption}
\usepackage{subcaption}
\usepackage[showframe=false]{geometry}
\usepackage{changepage}
\graphicspath{{images/}{dAcT/}}
\ifCLASSINFOpdf
\else
\fi
\begin{document}
%
\title{Mixed Gaussian-Impulse Noise Removal from Highly Corrupted Images via Adaptive Local and Nonlocal Statistical Priors}
%
%
%
\author{\IEEEauthorblockN{Nasser Eslahi, Hami Mahdavinataj, Ali Aghagolzadeh}\\
\IEEEauthorblockA{Department of Electrical and Computer Engineering\\
Babol University of Technology, Babol, Iran\\
Email: \{nasser.eslahi, mahdavinataj\}@stu.nit.ac.ir,  aghagol@nit.ac.ir}}
\maketitle

\begin{abstract}
The motivation of this paper is to introduce a novel framework for the restoration of images corrupted by mixed Gaussian-impulse noise. To this aim, first, an adaptive curvelet thresholding criterion is proposed which tries to adaptively remove the perturbations appeared during denoising process. Then, a new statistical regularization term, called joint adaptive statistical prior (JASP), is established which enforces both the local and nonlocal statistical consistencies, simultaneously, in a unified manner. Furthermore, a novel technique for mixed Gaussian plus impulse noise removal using JASP in a variational scheme is developed---we refer to it as De-JASP. To efficiently solve the above variational scheme, an efficient alternating minimization algorithm is developed based on split Bregman iterative framework. Extensive experimental results manifest the effectiveness of the proposed method comparing with the current state-of-the-art methods in mixed Gaussian-impulse noise removal.
\end{abstract}

\begin{IEEEkeywords}
Mixed Gaussian-impulse noise removal, Adaptive curvelet thresholding, nonlocal self-similarity.
\end{IEEEkeywords}

%

\section{Introduction}
Image denoising as a fundamental problem in image restoration has been extensively studied in image processing and computer vision (see [1]-[12] for few various representative works).
In this paper, we address the problem of image restoration from \emph{mixed Gaussian-impulse noise} data. Suppose the clean image $u\in \Bbb{R}^n$ is corrupted by both additive white Gaussian noise (AWGN) of $\eta\in \Bbb{R}^n$ (with variance $\sigma^2$) and an impulse noise. Also, let ${\Bbb{N}}_{imp}$ represent the process of image degradation with impulse noise. Accordingly, the overall degradation by mixed Gaussian-impulse noise can be modeled as the following:
\begin{equation}
f= {\Bbb{N}}_{imp}(u+\eta), 
\label{eq.1}
\end{equation}
where $f\in \Bbb{R}^n$ is the observed image (noisy measurements). In the literature, there are two common types of impulse noise: salt-and-pepper (SP) noise and random-valued (RV) noise. Let denote $u_{a,b}$ as the pixel value of image $u$ at location $(a,b)\in {\cal{I}}$, and $[d_{\min}, d_{\max}]$ be the dynamic range of $u$. The operator ${\Bbb{N}}_{imp} (\cdot)$  is defined as follows:
\begin{equation}
{\tilde{f}}_{a,b}={\Bbb{N}}_{imp}(u_{a,b})=\left \{\begin{array}{ccc} d_{a,b},& {\rm{with~probability}}~{{r}/{2}}, \\ d_{a,b}^{\prime}, &  {\rm{with~probability}}~{{r}/{2}}, \\ u_{a,b},& {\rm{with~probability}}~{1-r}, \end{array}\right.
\label{eq.2}
\end{equation}
where $d_{a,b}$ and $d_{a,b}^{\prime}$ are in the range of $[d_{\min}, d_{\max}]$ and independent from $u_{a,b}$. When $d_{a,b}=d_{\min}$ and $d_{a,b}^{\prime}=d_{\max}$, the model is SP noise; however, if $d_{a,b}=d_{a,b}^{\prime}$ is identically and uniformly distributed random numbers in $[d_{\min}, d_{\max}]$ with probability $r$, the model is called RV noise. Median-type filters, e.g. adaptive median filter (AMF) [13] and adaptive center-weighted median filter (ACWMF) [14], are widely used in the literature to remove impulse noise.

Let $\cal{N}$ denote the set of locations of noisy pixel candidates corrupted by impulse noise, and accordingly, the set of pixels that are likely to be uncorrupted by impulse noise is defined as $\cal{A}=\cal{I}/\cal{N}$. Therefore, the degradation matrix $\Phi$ can be defined as $\Phi_{a,b}:=\{1:(a,b)\in{\cal{A}}~or~0:(a,b)\in{\cal{N}}\}$ \footnote{In this paper, similar to [5], [7]-[12] and for the sake of a fair comparison, $\Phi$ is determined by AMF and ACWMF for SP and RV noise detection, respectively.}.
More precisely, the pixels probably being corrupted by impulse noise are assigned with $0$ while the rest of pixel values are more likely to be corrupted by AWGN. 
Delving further into this issue reveals this fact that, $\Phi\circ f$ follows an approximate Gaussian distribution, where $\circ$ denotes the element-wise multiplication between two matrices. 
Also, due to the \emph{ill-posed} nature of (1), regularization-based methods are often used to convert (1) into a \emph{well-posed} problem by imposing \emph{a priori} information of the underlying signal [7], [9]-[12].
Hence, from the Bayesian inference, the following regularization-based framework for mixed Gaussian-impulse noise removal is written as:
 \begin{align}
\mathop{\min }_{u} {1 \over 2}\|\Phi\circ(f-u)\|_{\ell_2}^2 +\lambda{\cal{J}}(u),
\label{eq.33}
\end{align}
where the first term is a penalty that represents the closeness of the solution to the observed scene; the second term is a regularization term which represents a priori information of the original scene; and $\lambda$ is a regularization parameter that balances the contribution of both terms. The term ${\cal{J}}(u)$ could be various choices, e.g., Tikhonov regularization [15], Geman and Reynolds' half quadratic variational model [16], Rudin, Osher and Fatemi's total variation model [17], Mumford-Shah model [18], and framelet based model [7].
These approaches do not need to find the damaged pixels and perform well in impulse noise removal. However, for the case of images corrupted by mixed Gaussian impulse noise, the Gaussian noise is not treated properly [12]. In recent works, the local smoothness and the nonlocal self-similarity properties (or image sparsity prior) exhibited in natural images are combined into the final cost functional of image restoration solution to achieve better performance, e.g., see [10]-[12].

The main contributions of this paper are listed as follows. First, we propose an \emph{adaptive curvelet thresholding} (ACT) criterion, in a statistical manner, for adaptively characterizing the perturbations appeared during noise removal process---exploiting local consistency. Second, we establish a new statistical regularization term, called \emph{joint adaptive statistical prior} (JASP) which enforces both the local and nonlocal statistical consistencies, simultaneously, in a unified manner. Third, we propose a novel technique for mixed Gaussian plus impulse noise removal using JASP in a variational scheme, we refer to it as De-JASP. To solve the above variational scheme, an efficient alternating minimization algorithm is developed based on split Bregman iterative framework [19].

The rest of this paper is organized as follows. Section II provides a brief background on a recently proposed modeling for nonlocal self-similarity. The proposed ACT is introduced in Section III. The proposed regularization term, along with its incorporation into the variational framework of mixed noise removal, and implementation details are introduced in Section IV. Numerical results are given in Section V and finally, Section VI concludes the paper.
\section{Background on Nonlocal Statistical Modeling (NLSM)}
Motivated by the success of BM3D [4] in image denoising, Zhang \emph{et al}. [11] proposed a model for nonlocal self-similarity prior information, which is employed efficiently in image restoration.
The NLSM explores the nonlocal self-similarity by means of the distribution of the transform coefficients, which are obtained by transforming the 3D array generated by stacking similar image patches.
To elucidate on, First, the image $u$ (of size $\sqrt{n}\times \sqrt{n}$) is divided into $P$ overlapped patches of equal sizes, i.e., $u_{p_{{l}^{\prime}}}$ (of size $\sqrt{B_p}\times \sqrt{B_p}$) at location ${l}^{\prime}$, where $l^{\prime}=1,2,\cdots,P$. Then, for each patch the best $c$ matched patches are found within a searching window (of size $\omega\times\omega$). Next, these similar patches are stacked into a 3D array, $G_{u_{p_{{l}^{\prime}}}}$, which is called a group. Now, a 3D transform, ${\cal{T}}^{3D}(\cdot)$, is conducted on the group to obtain its transform coefficients, followed by arranging them in a lexicographic order.
The mathematical formulation of the NLSM for self-similarity in transform domain can be written as:
\begin{equation}
\Psi_{{}_{\rm{NLSM}}}(u)=\|\Theta_u\|_{\ell_1}={\sum_{l^{\prime}=1}^{P}}\|{\cal{T}}^{3D}(G_{u_{p_{l^{\prime}}}})\|_{\ell_1},
\label{eq.patch}
\end{equation}
where $\Psi_{{}_{\rm{NLSM}}}(\cdot)$ corresponds to the NLSM operator.
To put it simply and briefly, after obtaining $\Theta_u$, the new estimate of $u$ is achieved by $\hat{u}=\Omega_{{}_{\rm{NLSM}}}(\Theta_u)$, where $\Omega_{{}_{\rm{NLSM}}}(\cdot)$ corresponds the inverse operator of $\Psi_{{}_{\rm{NLSM}}}$.
\section{The Proposed Adaptive Curvelet Thresholding (ACT)}
\begin{figure*}[t!]
{\hspace{-5.6cm}
         {\centering{
            \begin{subfigure}[b]{1\textwidth}
                 \includegraphics[scale=0.068]{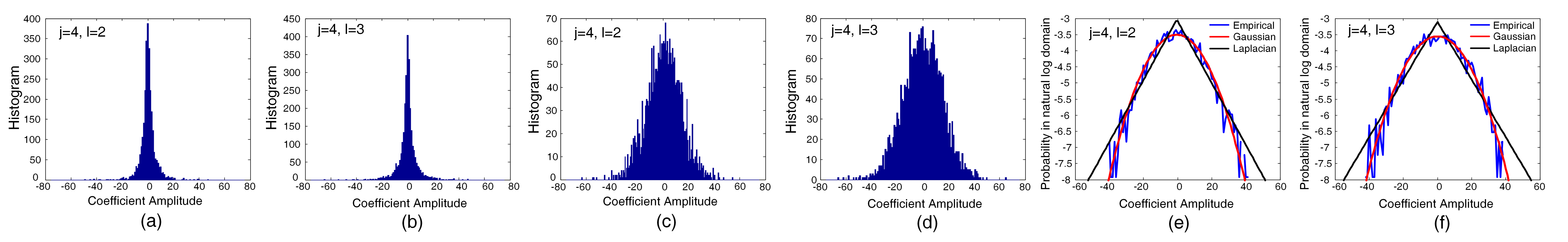}
                  \centering
         \end{subfigure}\\}}}%
         {\footnotesize{Fig. 1: (a)-(b) The histograms of two successive orientations at one scale of the curvelet coefficients of the clean image ``Fence" (which is shown in \emph{Fig. 2(a)}). (c)-(d) The histograms of the same orientations at the same scale of the curvelet coefficients of the initiated image $u_{init}$ (which is shown in \emph{Fig. 2(c)}), where, it is reconstructed by employing AMF on noisy image of \emph{Fig. 2(b)}. (e)-(f) PDFs of noise distribution (empirical perturbations) in natural log domain at the same orientations and scale (corresponding to the histograms of \emph{Figs. 1(a)-1(d)}) along with their best fitted PDFs: the noise distribution (blue), the Gaussian fit (red), and the Laplacian fit (black).}}
\end{figure*}%
The curvelet transform [20] is a multi-scale directional transform which provides a new multi-resolution representation with several features that are superior to existing representation, i.e., wavelets and steerable pyramids, thereby gaining good
performance in image denoising [1], [21]. For these reasons, here, an adaptive denoising/shrinkage procedure in curvelet domain via estimation of clean curvelet coefficients from available rough estimated data is proposed

We assume that the initially estimated image $u_{init}$ achieved by the median filter (AMF or ACWMF) is perturbed and model that perturbation by additive noise, with no prejudging on being AWGN or any specific species of noise. To elucidate on, this modeling of perturbations by additive noise helps us to model the noise probability density function (PDF) effectively and being able to solve the ensuing optimization problem efficiently.
Suppose, the initiated image coefficients in curvelet domain (${\hat{\vartheta}}=\Psi u_{init}$)\footnote{In matrix notation, we can write $\vartheta=\Psi u$, and $u=\Psi^{\dag}\vartheta$, where $\vartheta$ is the curvelet coefficient of function $u$, $\Psi$ is a forward curvelet transform and $\Psi^{\dag}=(\Psi^{T}\Psi)^{-1}\Psi^{T}$ is its Moore-Penrose pseudo-inverse transform (corresponding to the minimal dual synthesis frame). Due to the Parseval tight frame property of curvelet transform (i.e., $\Psi^{T}\Psi=I$), we have $\Psi^{\dag}=\Psi^{T}$.} can be written as: 
\begin{equation}
{\hat{\vartheta}}_{j,l,{\rm{\bf{k}}}}={\vartheta}_{j,l,{\rm{\bf{k}}}}+n_{j,l,{\rm{\bf{k}}}},
\label{eq.curvecoeff}
\end{equation}
where ${\hat{\vartheta}}$ and ${{\vartheta}}$ are the initiated image curvelet coefficient (available rough coefficient which is assumed to be noisy) and the noise-free (or clean) curvelet coefficients, respectively, and $n$ is unknown additive noise (or \emph{perturbation}) vector. Also, the three indices of $j,l,{\rm{\bf{k}}}$ are curvelet triple index of \emph{scale}, \emph{rotation} and \emph{position}, respectively. The problem of image denoising in curvelet domain can be expressed as estimation of clean coefficients from noisy data with Bayesian estimation techniques. Either minimum mean-squared error (MMSE) or maximum \emph{a posteriori} (MAP) estimator is used for solving this problem, the solution requires a prior knowledge about the distribution of curvelet coefficients. More specifically, these methods are optimized w.r.t. the marginal statistics of the coefficients within each sub-band, by imposing a prior distribution on the clean transform coefficients [6]. 

Figs. 1(a)-1(b) show the histograms of two successive orientations at one scale of the curvelet coefficients of the clean image ``Fence" ($256\times 256$)---see Fig. 2(a).
Obviously, it can be observed that the distributions are characterized by a very sharp peak at zero amplitude and extended tails to both sides of the peak (\emph{leptokurtic distribution}). This leptokurtic property is observed in all histograms of each scale and orientation of all test images.
In this paper, Laplacian distribution is chosen to model the marginal distributions of curvelet coefficients of the clean image, i.e., ${\Pr}_{{}_{{{\vartheta}}}}({{\vartheta}}_{j,l,{\rm{\bf{k}}}})=({{1}/{{\sqrt{2}}\sigma_{j,l,{\rm{\bf{k}}}}}})\cdot\exp{\big({{-{\sqrt{2}}~|{{\vartheta}}_{j,l,{\rm{\bf{k}}}}|}/{{\sigma_{j,l,{\rm{\bf{k}}}}}}}\big)}$,
where $\sigma_{j,l,{\rm{\bf{k}}}}$ is the standard deviation of clean coefficients.
This choice is despite the fact that marginal distributions of curvelet coefficients have significantly heavier tails than a Laplacian distribution; tend to go zero slower than a Gaussian distribution; and can be well modelled by a hyper-Laplacian distribution.
But, actually, this choice makes a trade-off between modeling the coefficients statistics accurately and being able to solve the ensuing optimization problem efficiently.

The histograms of the same two successive orientations at the same scale of the coefficients of the initially estimated image $u_{init}$---see Fig. 2 (c)---are shown in Figs. 1(c)-1(d). More specifically, these histograms are obtained using the curvelet coefficients of the initiated image $u_{init}$, which is achieved by the initial median filtering step. As can be seen, here, the leptokurtic behavior and heavy tails do not exist in the shown distributions. More accurately, these two properties do not exist in all distributions of each scale and orientation of all initiated images achieved by this step.

Considering (5), the noise distribution can be obtained using the curvelet coefficients of the clean image and the initiated image. Figs. 1(e)-1(f) show the PDFs of noise distribution at the same two successive orientations of the same scale corresponding to those of Figs. 1(a)-1(d). It can be observed that the empirical distributions of noise coefficients $n_{j,l,{\rm{\bf{k}}}}$ can be well characterized by Gaussian distributions, while the Laplacian distributions have much larger fitting errors.
This observation motivates us to model the PDF of $n_{j,l,{\rm{\bf{k}}}}$ by a standard Gaussian distribution: ${\Pr}_{{}_{n}}({n}_{j,l,{\rm{\bf{k}}}})=({{1}/{{\sqrt{2\pi}}\sigma_{{n}_{j,l}}}})\cdot\exp{\big({{-{n}_{j,l,{\rm{\bf{k}}}}^{2}}/{{2\sigma_{{n}_{j,l}}^{2}}}}\big)}$, where $\sigma_{{n}_{j,l}}^{2}$ is the noise variance in curvelet domain, which can be estimated using robust median estimator [22] and Monte-Carlo simulations [20].

\begin{figure}[t!]
{\hspace{1.2cm}
         {\centering
         \begin{subfigure}[b]{0.21\textwidth}
                 \includegraphics[scale=0.35]{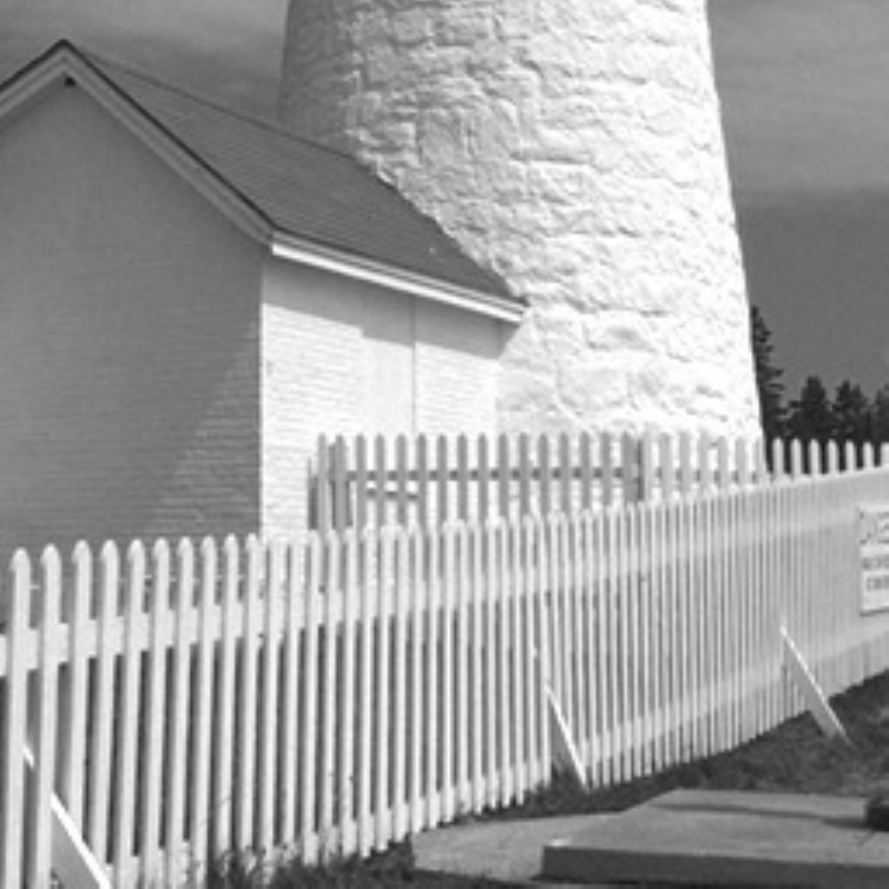}
                  \centering
                 \caption{}
                 \label{fig:1}
         \end{subfigure}%
         \begin{subfigure}[b]{0.201\textwidth}
                 \includegraphics[scale=0.35]{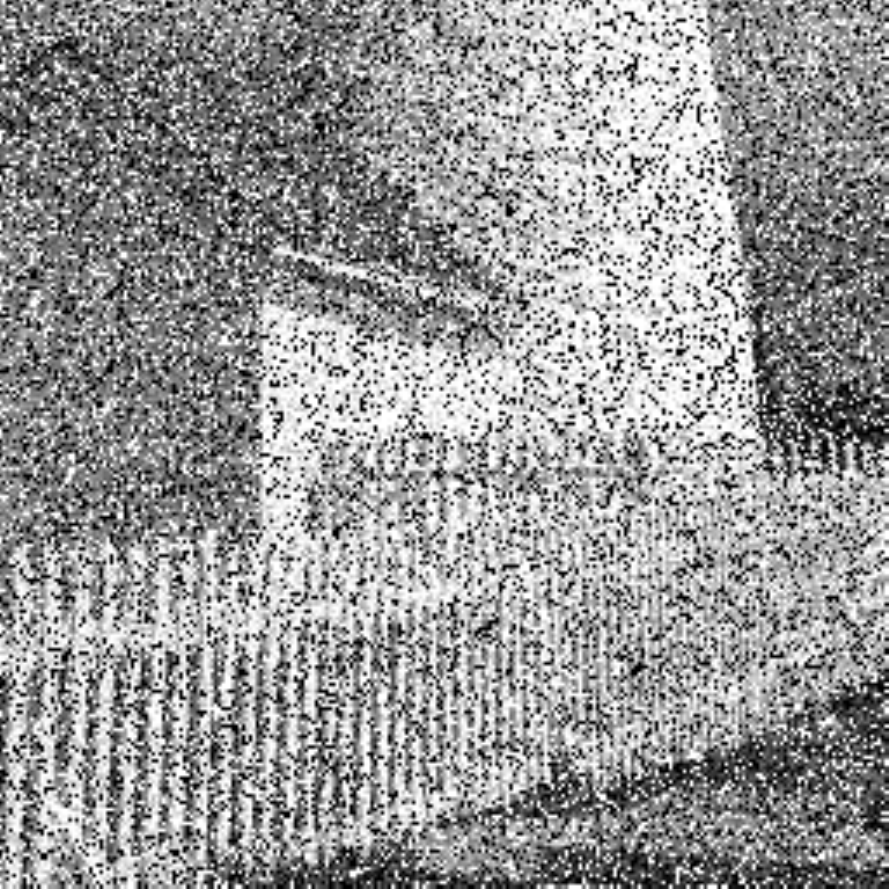}
                 \centering
                 \caption{}
                 \label{fig:1}
         \end{subfigure}
         \begin{subfigure}[b]{0.201\textwidth}
                 \includegraphics[scale=0.35]{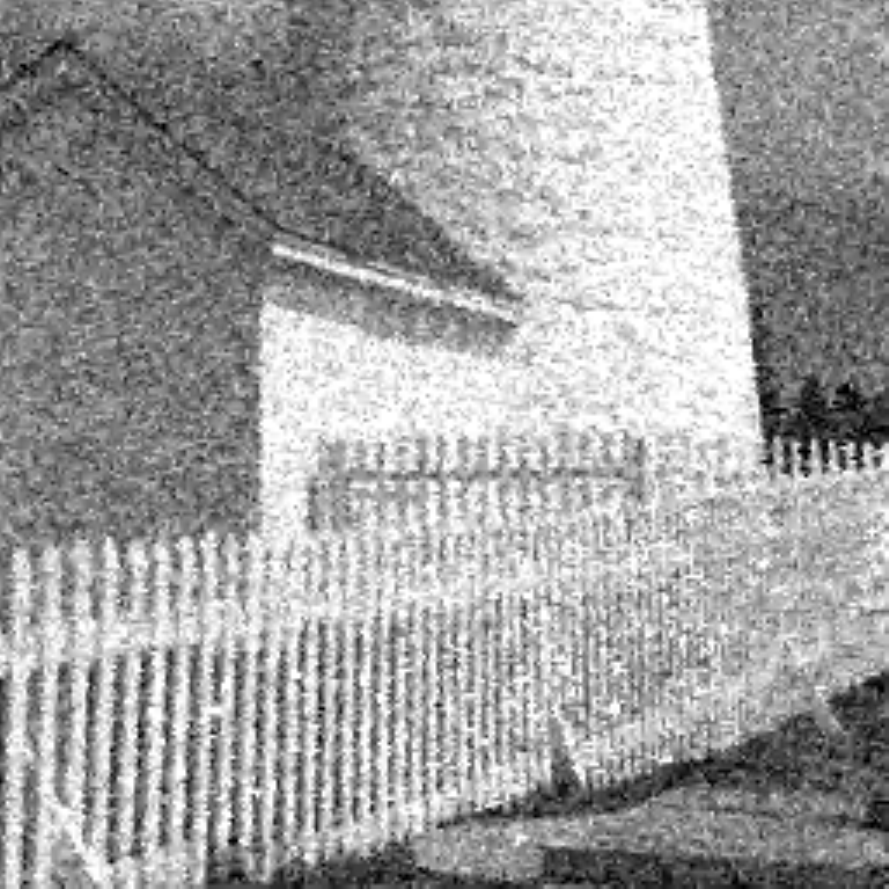}
                  \centering
                 \caption{}
                 \label{fig:1}
         \end{subfigure}%
         \begin{subfigure}[b]{0.21\textwidth}
                 \includegraphics[scale=0.35]{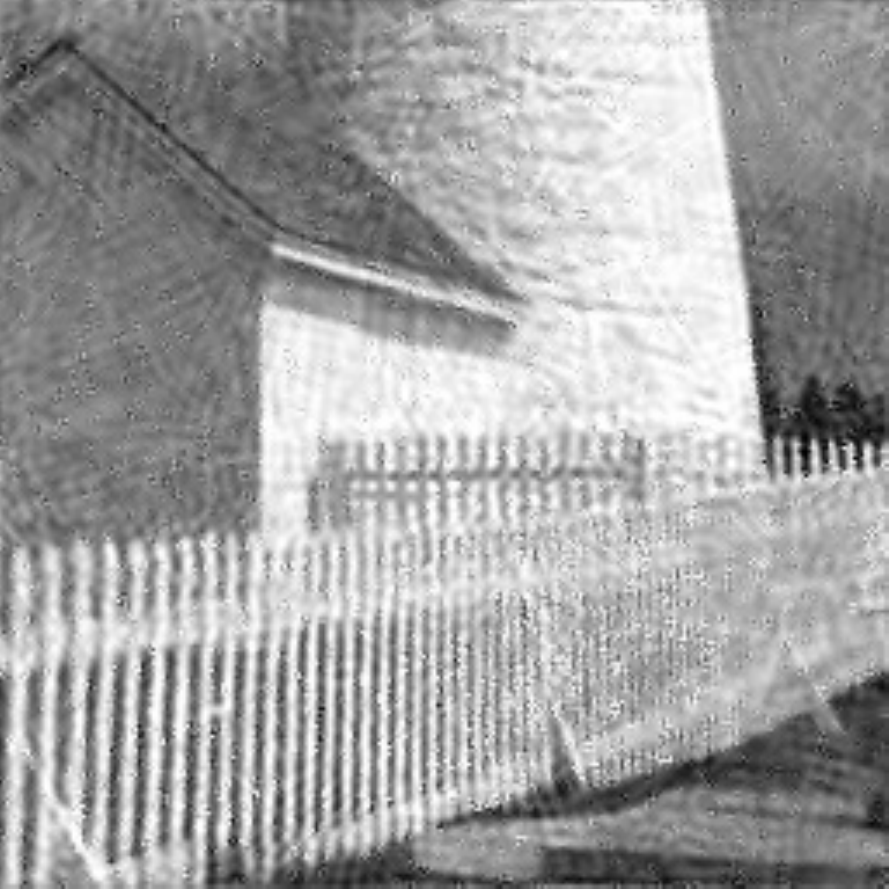}
                 \centering
                 \caption{}
                 \label{fig:1}
         \end{subfigure}}}\\
{\footnotesize{Fig. 2: (a) Original \emph{Fence} image (of szie $256\times 256$). (b) The noisy image which is corrupted by Gaussian plus salt-and-pepper impulse noise with $\sigma=30$ and $r_{sp}=30\%$ (PSNR=9.92 dB, SSIM=0.143). (c) The initiated image using AMF (PSNR=18.81 dB, SSIM=0.432). (d) The recovered image via ACT by 11 iterations (PSNR=22.11 dB, SSIM=0.589).}}
\end{figure}

Consider the marginal distribution of the clean curvelet coefficients. Since in image denoising problems, we do not have access to the clean image, the variance of the clean image curvelet coefficients ${{\sigma_{{j,l,{\rm{\bf{k}}}}}^{2}}}$ can be estimated by \emph{maximum likelihood} (ML) estimator, through averaging over the neighboring initiated image coefficients (available ${\hat{\vartheta}}_{j,l,{\rm{\bf{k}}}}$)  within a squared window [23].

In order to estimate ${{\vartheta}_{j,l,{\rm{\bf{k}}}}}$ from the available observation ${\hat{\vartheta}_{j,l,{\rm{\bf{k}}}}}$, the MAP estimator is employed:
\begin{equation}
{\vartheta}_{j,l,{\rm{\bf{k}}}}=\arg\max_{{\vartheta}_{j,l,{\rm{\bf{k}}}}}\bigg\{{{\Pr}_{{}_{{{\vartheta}}|\hat\vartheta}}}\big({\vartheta_{j,l,{\rm{\bf{k}}}}}|{{\hat{\vartheta}}_{j,l,{\rm{\bf{k}}}}}\big)\bigg\}.
\label{eq.map}
\end{equation}
By considering (5) and using Bayes' rule, we get:
\begin{align}
{\vartheta}_{j,l,{\rm{\bf{k}}}}=\arg\max_{{\vartheta}_{j,l,{\rm{\bf{k}}}}}\bigg\{{{\Pr}_{{}_{{\hat{\vartheta}}|\vartheta}}}\big({\hat\vartheta_{j,l,{\rm{\bf{k}}}}}|{\vartheta_{j,l,{\rm{\bf{k}}}}}\big)\cdot{{\Pr}_{{}_{{\vartheta}}}\big({\vartheta_{j,l,{\rm{\bf{k}}}}}\big)}\bigg\} \nonumber \\
=\arg\max_{{\vartheta}_{j,l,{\rm{\bf{k}}}}}\bigg\{{{\Pr}_{{}_{n}}}\big({\hat\vartheta_{j,l,{\rm{\bf{k}}}}}-{\vartheta_{j,l,{\rm{\bf{k}}}}}\big)\cdot{{\Pr}_{{}_{{\vartheta}}}\big({\vartheta_{j,l,{\rm{\bf{k}}}}}\big)}\bigg\}.
\label{eq.bays}
\end{align}
Therefore, (7) allows us to write this estimation in terms of the PDF of the noise coefficient ($\Pr_{{}_{n}}$), and the PDF of the clean image coefficient ($\Pr_{{}_{\vartheta}}$).
By substituting $\Pr_{{}_{n}}$ and $\Pr_{{}_{\vartheta}}$ into (7) and doing some manipulations, the estimation of $\vartheta_{j,l,{\rm{\bf{k}}}}$ is achieved as:
\begin{equation}
{\vartheta}_{j,l,{\rm{\bf{k}}}}={\rm{sign}}(\hat\vartheta_{j,l,{\rm{\bf{k}}}})\max\big({0,|\hat\vartheta_{j,l,{\rm{\bf{k}}}}|-{\frac{\sqrt{2} \sigma_{{n}_{j,l}}^{2}}{\sigma_{j,l,{\rm{\bf{k}}}}}}}\big),
\label{eq.varshrink}
\end{equation}
where, in fact, (8) is the classical soft shrinkage function [24] defined as ${\cal{S}}_{\rho}(y):={\rm{sgn}}(y)\cdot \max(0, |y|-\rho)$.
Not only can the proposed ACT applied for the initiated image achieved by the initial median filtering step, but also it can be employed in further iterations---by using the last approximated image achieved in previous iteration as the noisy image for the current iteration---which leads to a simple and effective mixed Gaussian-impulse noise removal scheme, e.g., see Fig. 2(d).

\section{The Proposed Image Denoising via JASP (De-JASP)}
The proposed \emph{joint adaptive statistical prior} (JASP) is defined by integrating both the local information prior (which depicts the local smoothness and geometric regularity of image structures achieved by discrete curvelet transform (DCuT) [20]) and the nonlocal self-similarity prior (corresponding to the NLSM in 3D transform domain achieved by the method introduced in Section II). In a mathematical expression, the proposed JASP is expressed as:
\begin{align}
{\cal{J}}_{{}_{\rm{JASP}}}(u)&=\Psi_{{}_{\rm{DCuT}}}(u)+{{\tau}}\Psi_{{}_{\rm{NLSM}}}(u) \nonumber \\
&=\|\Psi u\|_{\ell_1}+{{\tau}}\|\Theta_u\|_{\ell_1},
\label{eq.jasr}
\end{align}
where the first and the second terms represent the image local smoothness prior and nonlocal self-similarity prior, respectively. Also, ${{\tau}}$ is a regularization parameter which controls the trade-off between two competing (statistical) terms.

By substituting the proposed JASP (9) for the regularization term ${\cal{J}}(u)$ in the regularization-based framework of (3), and by introducing two auxiliary variables $\vartheta$ and $w$, the proposed optimization problem for image recovery is expressed as follow:
 \begin{align}
\mathop{\min }_{u,\vartheta, w} \{{1 \over 2}\|\Phi\circ(f-u)\|_{\ell_2}^2 +\lambda\|\vartheta\|_{\ell_1}+{\tau}^{\prime} \|\Theta_w\|_{\ell_1}\} \nonumber \\
s.t.~~~~\vartheta=\Psi u,~ u=w,~ w=\Omega_{{}_{\rm{NLSM}}}(\Theta_w).
\label{eq.constrained}
\end{align}
In order to solve the above minimization problem, an alternating split Bregman iterative algorithm [19] is invoked. We finally obtain the following schemes:
   \begin{subequations}\label{eq:subeqns}
     \begin{align}
(&u^{k+1}, \vartheta^{k+1}, w^{k+1})\leftarrow\arg\min_{u, \vartheta, w}\Big\{{{1\over 2}{\|\Phi\circ(f-u)\|_{\ell_2}^2}+{\lambda}\|\vartheta\|_{\ell_1}}\nonumber\\&{+{\tau}^{\prime} \|\Theta_w\|_{\ell_1}+{\frac{\mu_1}{2}}{\|\vartheta-\Psi u-b^{k}\|_{\ell_{2}}^2}}
+{\frac{\mu_2}{2}}{\|u-w-c^{k}\|_{\ell_{2}}^2}\Big\} \label{eq:subeq1}\\
&b^{k+1}\leftarrow b^{k}-\vartheta^{k+1}+\Psi u^{k+1}\label{eq:subeq4}\\
&c^{k+1}\leftarrow c^{k}-u^{k+1}+w^{k+1}.\label{eq:subeq5}
     \end{align}
    \end{subequations}
Here, $\mu_1$ and $\mu_2$ are fixed value parameters for improving the numerical stability of the algorithm.

Given $\vartheta^{k}$ and $w^k$, the $u$ \emph{sub-problem} of (11a) consists of minimizing a strictly convex quadratic function that can be solved easily, i.e., $u^{k+1}={\big(\Phi^{T}\Phi+\mu I\big)}^{-1}\circ {\cal{D}}$, where $\mu=\mu_1+\mu_2$ and ${\cal{D}}=\big(\mu_1\Psi^{T}(\vartheta^{k}-b^{k})+\mu_2(w^{k}+c^{k})+\Phi^{T}\Phi\circ f\big)$. In order to avoid computing the matrix inversion in the first term, the matrix inversion lemma of the Sherman-Morrison formula is applied to that term. Thus, the $u$ \emph{sub-problem} can be obtained as:
\begin{equation}
u^{k+1}={\frac{1}{\mu}}\big({I-\frac{\Phi^{T}\Phi}{\mu+\Phi\Phi^{T}}}\big)\circ {\cal{D}}.
\label{eq.usubp}
\end{equation}

By given $u$ in hand (according to (12)) and considering $\hat\vartheta=\Psi u+b$ (for simplicity, the superscript $k$ is dropped without confusion), the $\vartheta$ \emph{sub-problem} of (11a) becomes:
\begin{equation}
\min_{\vartheta}{{1\over 2}{\|\vartheta-\hat\vartheta\|_{\ell_{2}}^{2}+{\lambda \over \mu_1}{\|\vartheta\|_{\ell_{1}}}}}.
\label{eq.coeffsubp}
\end{equation}
By these transformations, we regard $\vartheta^{\prime}$ as some type of the noisy unknown coefficient $\vartheta$. More precisely, the sub-problem of (13) can be interpreted as the denoising in curvelet coefficient domain.
Considering the fact that the unknown variable $\vartheta$ is component-wise separable in curvelet domain, each of its component ${\vartheta}_{j,l,{\rm{\bf{k}}}}$ can be obtained by a component-wise shrinkage procedure:
\begin{equation}
{\vartheta}_{j,l,{\rm{\bf{k}}}}=\arg\min_{{\vartheta}_{j,l,{\rm{\bf{k}}}}}\bigg\{{\frac{1}{2}}{\big|}{\vartheta}_{j,l,{\rm{\bf{k}}}}-{\hat\vartheta}_{j,l,{\rm{\bf{k}}}}{\big|}^{2}+{\frac{\lambda_{j,l,{\rm{\bf{k}}}}}{\mu_1}}{\big|}{\vartheta}_{j,l,{\rm{\bf{k}}}}{\big|}\bigg\},
\label{eq.cws}
\end{equation}
where its solution is the same as the one introduced in (8) with regarding ${\frac{\lambda_{j,l,{\rm{\bf{k}}}}}{\mu_1}}={\frac{{\sqrt{2}}\sigma_{n_{j,l}}^{2}}{\sigma_{j,l,{\rm{\bf{k}}}}}}$. To elucidate on, the $\vartheta$ \emph{sub-problem} of (13) can be interpreted as the proposed ACT introduced in Section III. Thus, the parameter $\lambda$ is self-adaptive.

Given $u$, analogously, the $w$ \emph{sub-problem} of (11a) can be written as:
\begin{equation}
\min_{w}{{1\over 2}{\|w-z\|_{\ell_{2}}^{2}+{{\tau}^{\prime} \over \mu_2}{\|\Theta_w\|_{\ell_{1}}}}},
\label{eq.wsubp}
\end{equation}
where $z=u-c$ (for simplicity the superscript $k$ is omitted without confusion). Owing to the complicated definition of $\Theta_w$, it seems difficult to solve (15) directly. In order to solve (15) amenable, according to [11], a reasonable assumption is used which leads to obtain a closed-form solution of (15). By assuming that all elements of $w-z$ (such that $w, z\in \Bbb{R}^{n}$) are independent and identically distributed (i.i.d.) with zero-mean and variance ${\sigma^{\prime}}^{2}$ \footnote{It is worth emphasizing that the above assumption does not need to be Gaussian, Laplacian or generalized Gaussian process.}, and by invoking the \emph{law of large numbers} in probability theory, for any $\epsilon^{\prime}>0$, it leads to:
\begin{equation}
\lim_{n\rightarrow\infty}\Pr\{{\big|}{{1}\over{n}}\|w-z\|_{\ell_{2}}^{2}-{\sigma^{\prime}}^{2}{\big|}<{{\epsilon^{\prime}}\over{2}}\}=1.
\label{eq.lln1}
\end{equation}
\begin{table}[!t]
\renewcommand{\arraystretch}{1.2}
{\footnotesize{{~~~~~~~~~~~~~~~~~TABLE I: T\sc he Proposed De-JASP Framework}}}\\\\
\label{table_example}
\centering
\begin{tabular}{l}
\hline\hline
 {\bfseries{Input}}: $f$, $B_p, \omega, {\tau}^{\prime}, \mu_1, \mu_2, Tol$, $k_{max}^{\prime}$: maximum iteration number\\ 
 {\bfseries{Output}}: $u^*$: Recovered image;\\
 {\bfseries{Initialization}}: Set $k=0$; $(b^{0},c^{0})=(0,0)$\\
$u_{init}=u^{0}$= median-filtering$(f)$; compute $\Phi$ \\
 {\bfseries{While}} a stop criterion is not satisfied  {\bfseries{do}}\\
1.~~~~~$u^{k+1}=\tilde{u}=u^{k}$;~${\hat{\vartheta}}^{k+1}=\Psi u^{k+1}+b^{k}$;~$z^{k+1}=u^{k+1}-c^{k}$\\
2.~~~~~solve $\vartheta$ \emph{sub-problem} using (14) and concatenating all coefficients\\
3.~~~~~solve $w$ \emph{sub-problem} using (19)\\
4.~~~~~solve $u$ \emph{sub-problem} using (12)\\
5.~~~~~compute $s^{k+1}$= SSIM($u^{k+1}, \tilde{u}$)\\
6.~~~~~compute \emph{diff} = {\rm {abs}}$(s^{k+1}-s^{k})$\\
7.~~~~~update $b^{k}, c^{k}$ using (11b) and (11c), respectively\\
8.~~~~~$k\leftarrow k+1$\\
{\bfseries{end}} While\\
\hline
{\bfseries{stopping criterion}}: $k=k_{max}^{\prime}$ or {\emph{diff}} $\leq Tol$$$.\\ \hline 
\end{tabular}
\end{table}
Due to the orthogonal property of 3D transform ${\cal{T}}^{3D}$, and denoting the error vector in 3D transform domain by $\Theta_e=\Theta_w-\Theta_z$ (such that $\Theta_w, \Theta_z\in \Bbb{R}^{K_{\Theta}}$, where ${K_{\Theta}}=B_p*c*P$), it is concluded that all elements of $\Theta_e$ are i.i.d. with zero-mean and variance ${\sigma^{\prime}}^{2}$. Thanks to the \emph{law of large numbers}, by doing the same manipulations with (16) to $\Theta_e$, for any $\epsilon^{\prime}>0$, it yields:
\begin{equation}
\lim_{K_{\Theta}\rightarrow\infty}\Pr\{{\big|}{{1}\over{K_{\Theta}}}\|\Theta_w-\Theta_z\|_{\ell_{2}}^{2}-{\sigma^{\prime}}^{2}{\big|}<{{\epsilon^{\prime}}\over{2}}\}=1.
\label{eq.lln2}
\end{equation}
Therefore, according to (16) and (17) the following relationship is described (almost surely): ${\frac{1}{n}}\|w-z\|_{\ell_{2}}^{2}={\frac{1}{K_\Theta}}\|\Theta_w-\Theta_z\|_{\ell_{2}}^{2}$, where incorporating it into (15) leads to:
\begin{equation}
\arg\min_{\Theta_w}{{1\over 2}{\|\Theta_w-\Theta_z\|_{\ell_{2}}^{2}+{\frac{K_{\Theta} {\tau}^{\prime}}{\mu_2~n}}{\|\Theta_w\|_{\ell_{1}}}}}.
\label{eq.newwsub}
\end{equation}
Since the unknown variable $\Theta_w$ is component-wise separable in (18), each of its component $\Theta_w(j^{\prime}), j^{\prime}=1,\cdots,K_{\Theta}$ can be independently obtained by a component-wise (soft) shrinkage procedure. Thus, the closed-form solution of (15) is written as:
\begin{equation}
w=\Omega_{{}_{\rm{NLSM}}}\big({{\Theta_w}}\big)=\Omega_{{}_{\rm{NLSM}}}\big({{{\cal{S}}_{{\frac{K_{\Theta} {\tau}^{\prime}}{\mu_2n}}}\big(\Theta_z\big)}}\big).
\label{eq.omega}
\end{equation}

The proposed De-JASP algorithm is delineated in Table I.

\section{Experimental Results}
In this section, we evaluate the performance of the proposed method---De-JASP.
The performances of our experiments are evaluated on 4 gray-scale test images of \emph{Barbara}, \emph{Boat}, \emph{Fence}, and \emph{Parrot} (of size $256\times 256$). To evaluate our simulation results, we use two applicable quality assessors, the pick signal-to-noise ratio (PSNR) in dB and the structural similarity (SSIM) [25].
All the experimental results reported in this paper are averaged over 10 independent trials. Also, the second generation of DCuT via wrapping [20] is employed in our implementations. To demonstrate the effectiveness of the proposed method in image denoising, we have compared it with two competitive mixed noise removal techniques: (i) iterative framelet-based approximation/sparsity deblurring algorithm IFASDA [7], and (ii) image denoising via joint statistical modeling (JSM) [11].
It is worth emphasizing that these two recently proposed denoising methods are among the state-of-the-art techniques  in mixed-noise removal which achieve the best performance so far.
Note that, to make a fair comparison among the competing methods, we have carefully tuned their parameters to achieve the best performance. Also, for the sake of fair comparison, the same test conditions are used in all experiments; i.e., the same noisy measurements are applied for each method. Due to the space limitations, only parts of the experimental results are shown in this paper. Interested readers may contact the corresponding author for all the images and the source code.

In the following, the effectiveness of De-JASP is investigated. In our implementation, in process of finding NLSM for self-similarity in 3D transform domain, the size of each patch is set to $8\times 8$. The size of training window for searching matched blocks is set to $20\times 20$, and the number of best matched blocks is set to $10$. The other main parameters of the proposed algorithm are empirically set: $k_{max}^{\prime}=20$, $Tol=10^{-3}$, $\mu_1=0.16\mu$, $\mu_2=0.84\mu$ and $\mu=2.7\times 10^{-3}$. Also, the orthogonal 3D transform denoted by ${\cal{T}}^{3D}$ is composed of 2D DCT and 1D Haar transform.
\begin{table*}[!t]
\renewcommand{\arraystretch}{1}
{\footnotesize{\footnotesize{~~~~~~~~~~~~~~~~~TABLE II:~\sc{PSNR/SSIM Comparisons for Gaussian plus Salt-and-Pepper Noise Removal.}}}}\\~\\
\label{table_example}
\centering
\begin{tabular}{|c|c|c|c|c|c|c|c|}\cline{6-8}
\multicolumn{5}{c|}{}&\multicolumn{3}{c|}{Algorithm}\\ \cline{6-8}\hline 
Test Image&$r_{sp}$&$\sigma$&Noisy Image& Init. Image&IFASDA [7]&JSM [11]& De-JASP\\ \hline
{}&{0.2}& 20&12.10/0.147&22.48/0.529 &24.92/0.756&27.73/0.795&\bf{28.06/0.835}\\
{Barbara}&{0.2} &30&11.69/0.133 &19.57/0.385& 20.46/0.514&22.76/0.537&\bf{24.83/0.676}\\
{}&{0.3} &30& 10.21/0.091& 19.10/0.364& 20.34/0.502& 22.63/0.533 &\bf{24.43/0.655}\\ \hline\hline
{}&{0.2}& 20&12.14/0.134 &23.08/0.510 &27.81/0.771 & 28.40/0.784&\bf{29.22/0.833}\\
{Boat}&{0.2} & 30 &11.72/0.122 &19.93/0.373 & 22.74/0.506&23.22/0.520 &\bf{25.88/0.679}\\
{}&{0.3}& 30& 10.24/0.085& 19.47/0.355&22.67/0.499 & 23.15/0.518&\bf{25.47/0.659}\\ \hline\hline
{}&{0.2}& 20&11.77/0.212 & 21.58/0.560&24.99/0.782 & 26.60/0.791&\bf{26.79/0.812}\\
{Fence}&{0.2} & 30 &11.39/0.197 &19.13/0.443 & 21.03/0.577&22.35/0.583 &\bf{24.15/0.693}\\
{}&{0.3}& 30& 9.92/0.143& 18.81/0.432& 20.82/0.569&22.16/0.577 &\bf{23.80/0.679}\\ \hline\hline
{}&{0.2}& 20&11.89/0.095 & 23.15/0.407& 27.78/0.759& 28.83/0.765&\bf{29.65/0.837}\\
{Parrot}&{0.2} & 30 &11.50/0.087 &19.97/0.277 & 22.65/0.424& 23.51/0.438&\bf{26.21/0.640}\\
{}&{0.3}& 30& 10.02/0.063&19.51/0.264 &22.46/0.411 &23.31/0.435 & \bf{25.61/0.616}\\ \hline
\end{tabular}
\end{table*}

Table II lists the PSNR and SSIM values of the recovered noisy test images obtained by the proposed De-JASP compared to those obtained by JSM and IFASDA in different noise level hypotheses (best results are emphasized in bold face). From Table II, it can be inferred that the performance of the proposed De-JASP is superior to those of compared algorithms in terms of both employed objective quality assessors. The average PSNR (SSIM) gain of De-JASP over JSM and IFASDA, in mixed \emph{AWGN+RV} noise removal, can be up to 2.31 dB (0.154) and 3.55 dB (0.167), respectively.
For visual comparison, some recovered images by the competing methods, in different noise levels, are shown in Figs. 4-7. It can be observed that De-JASP outperforms the other competing methods and reproducing clearer images. Better visual comparison can be made by zooming the images on the screen.

\begin{figure}
{\hspace{1.2cm}
         {\centering
         \begin{subfigure}[b]{0.21\textwidth}
                 \includegraphics[scale=0.35]{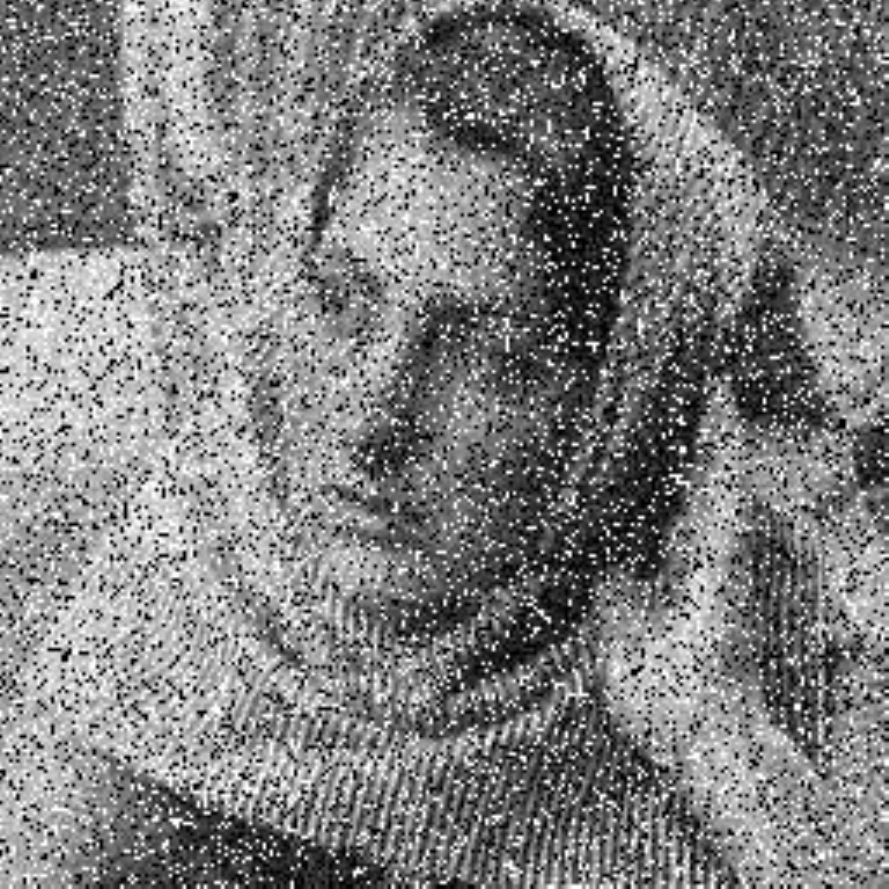}
                  \centering
                 \caption{}
                 \label{fig:1}
         \end{subfigure}%
         \begin{subfigure}[b]{0.201\textwidth}
                 \includegraphics[scale=0.35]{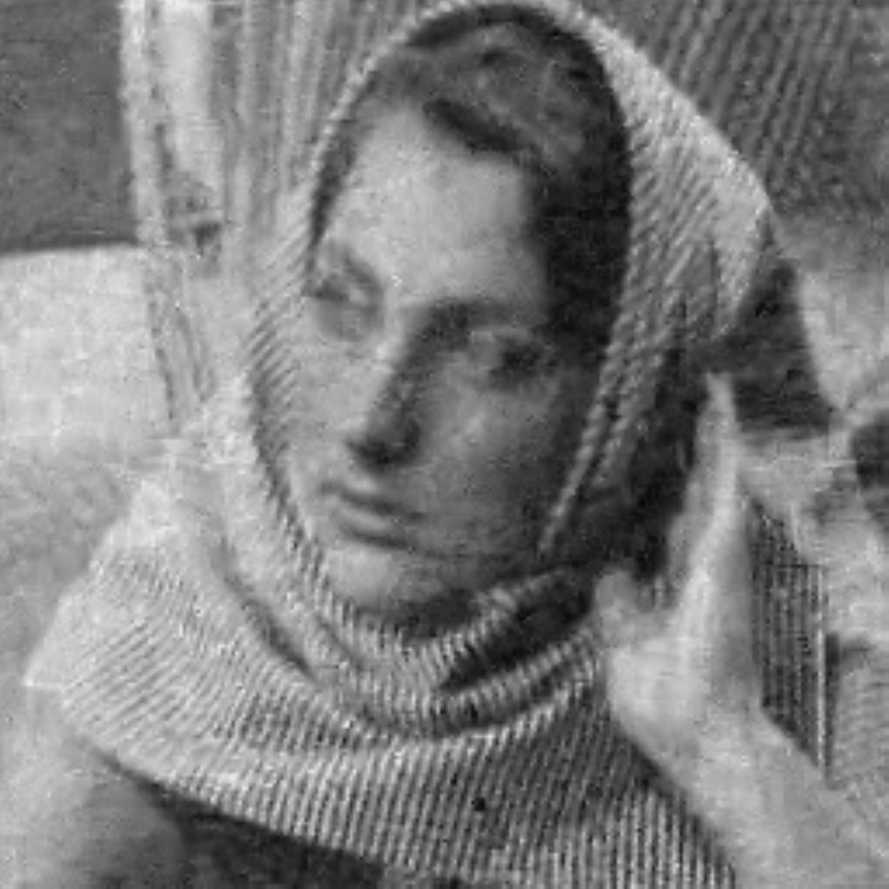}
                 \centering
                 \caption{}
                 \label{fig:1}
         \end{subfigure}
         \begin{subfigure}[b]{0.201\textwidth}
                 \includegraphics[scale=0.35]{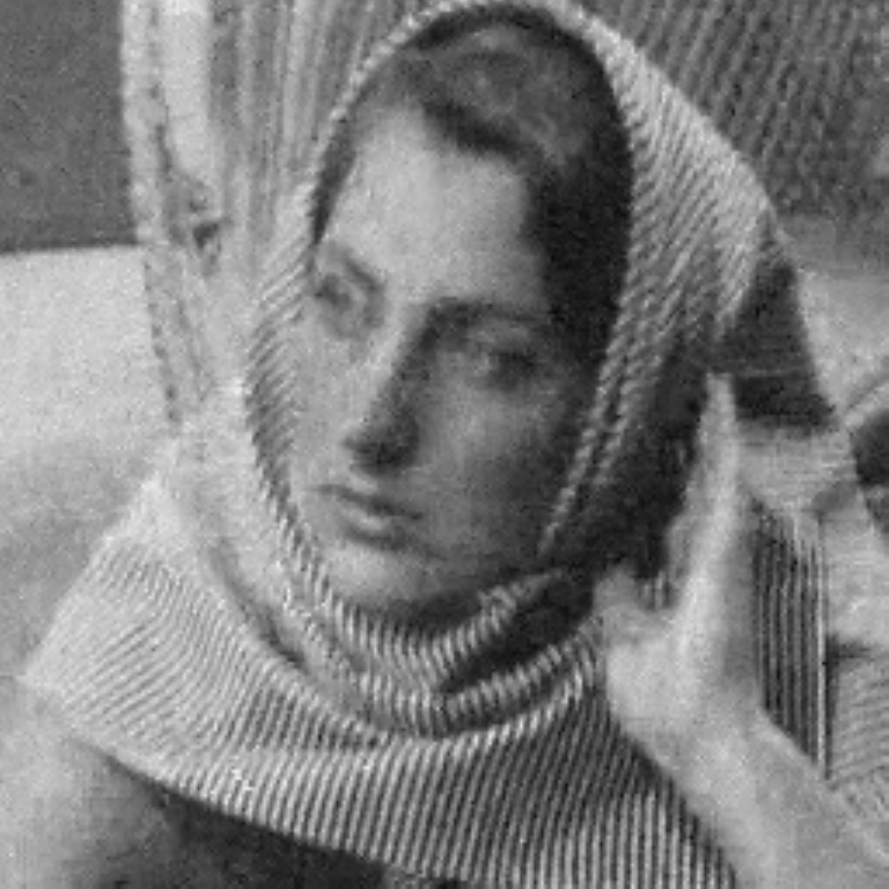}
                  \centering
                 \caption{}
                 \label{fig:1}
         \end{subfigure}%
         \begin{subfigure}[b]{0.21\textwidth}
                 \includegraphics[scale=0.35]{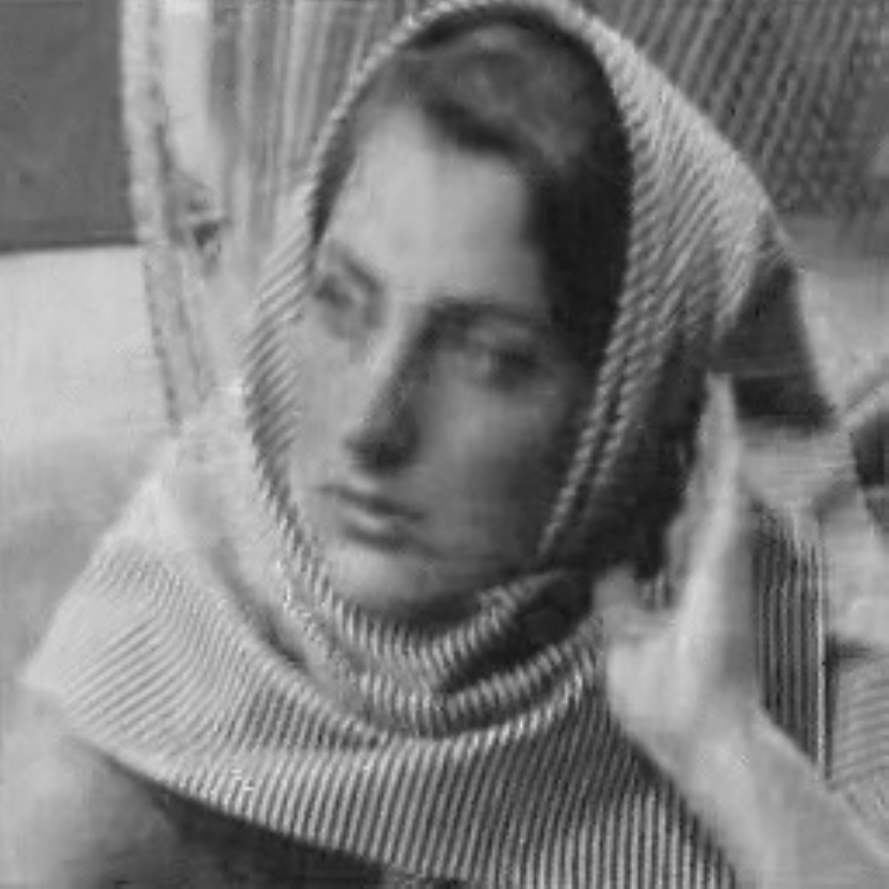}
                 \centering
                 \caption{}
                 \label{fig:1}
         \end{subfigure}}}\\
{\footnotesize{Fig. 4: Visual quality comparison of mixed Gaussian-impulse noise removal on image ``Barbara". (a) noisy image corrupted by AWGN+SP noise with $\sigma=20$ and $r_{sp}=20\%$. Denoised results by (b) IFASDA [7]; (c) JSM [11]; and (d) the proposed De-JASP. For numerical comparison, see Table II.}}
\end{figure}
\begin{figure}
{\hspace{1.2cm}
         {\centering
         \begin{subfigure}[b]{0.21\textwidth}
                 \includegraphics[scale=0.35]{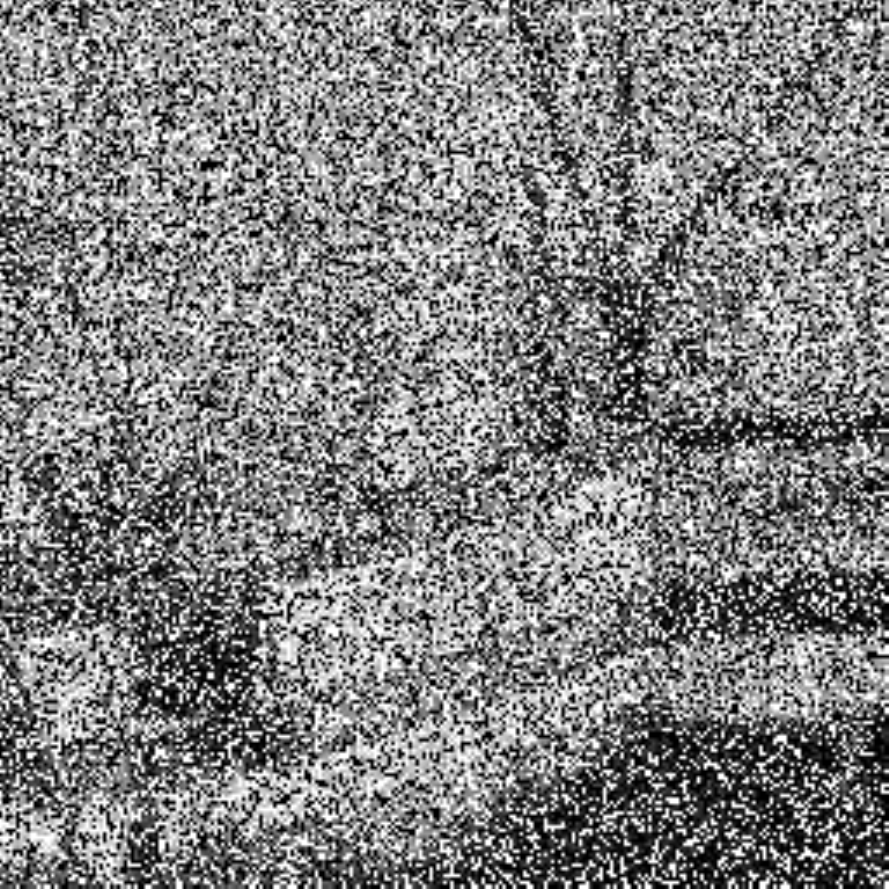}
                  \centering
                 \caption{}
                 \label{fig:1}
         \end{subfigure}%
         \begin{subfigure}[b]{0.201\textwidth}
                 \includegraphics[scale=0.35]{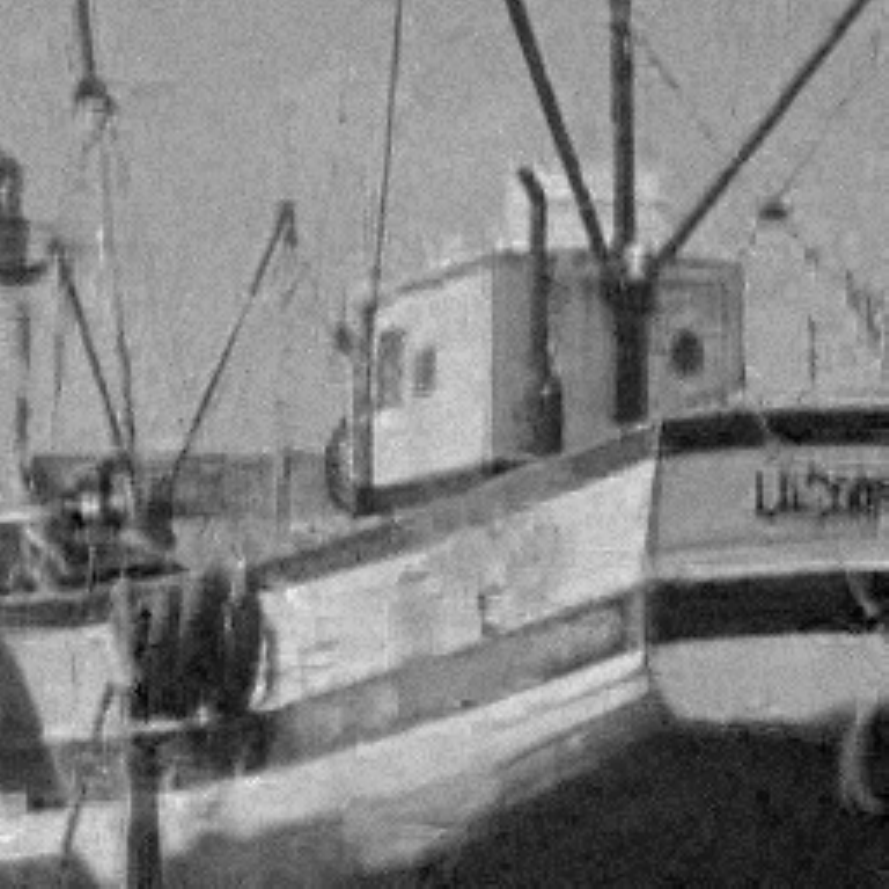}
                 \centering
                 \caption{}
                 \label{fig:1}
         \end{subfigure}
         \begin{subfigure}[b]{0.201\textwidth}
                 \includegraphics[scale=0.35]{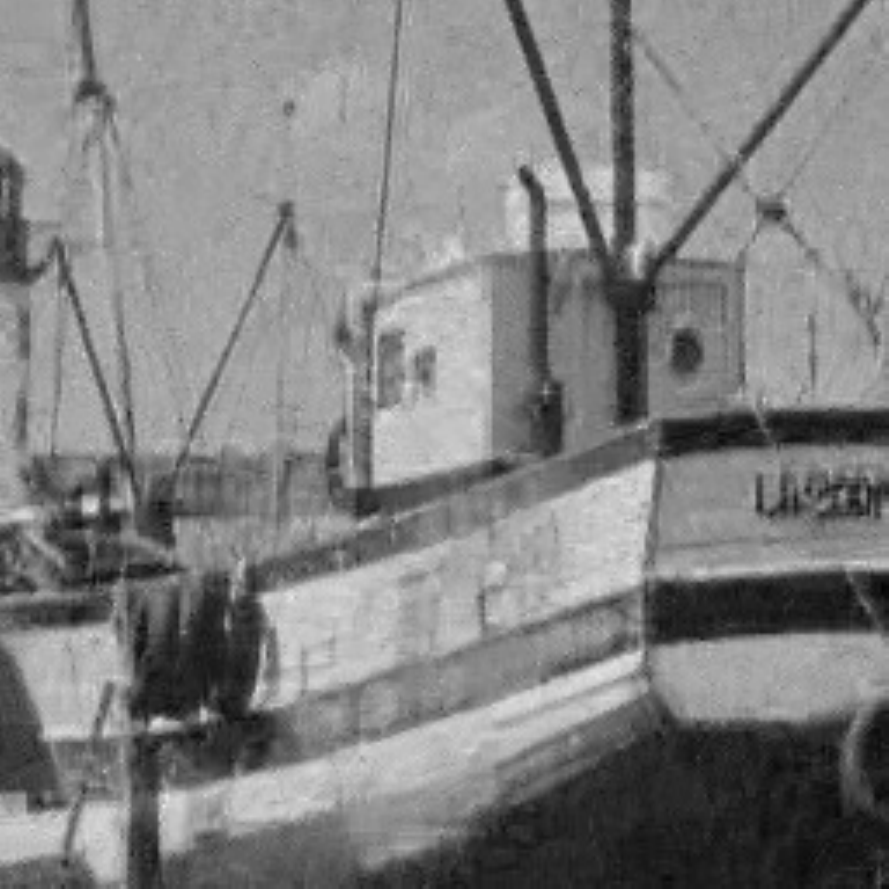}
                  \centering
                 \caption{}
                 \label{fig:1}
         \end{subfigure}%
         \begin{subfigure}[b]{0.21\textwidth}
                 \includegraphics[scale=0.35]{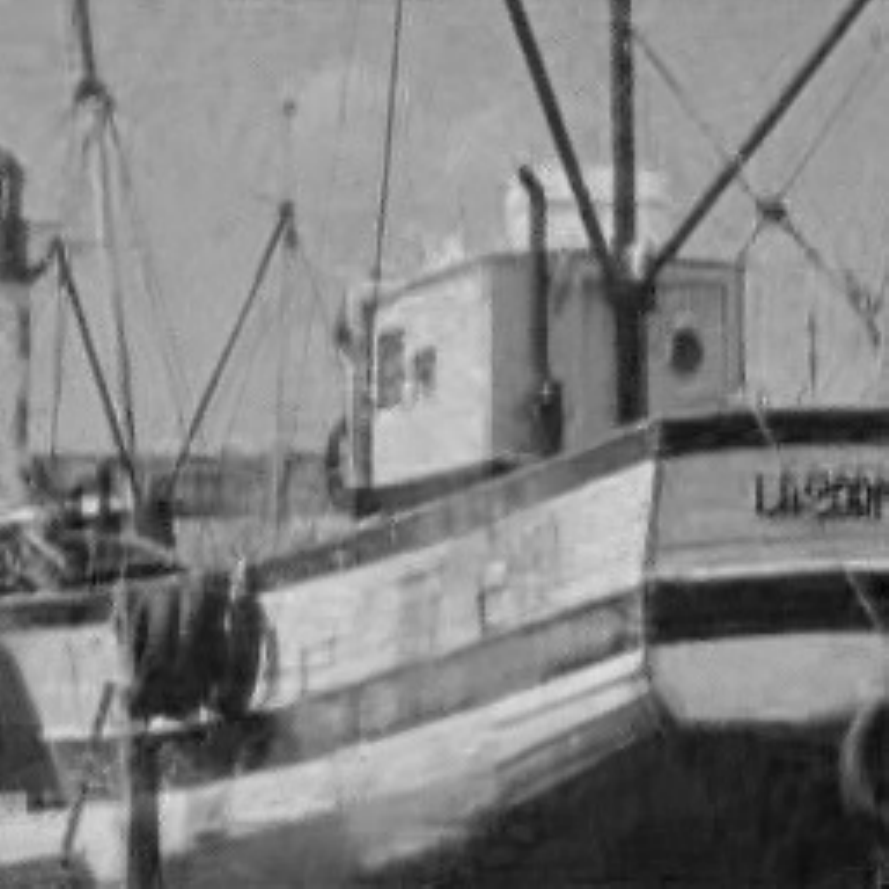}
                 \centering
                 \caption{}
                 \label{fig:1}
         \end{subfigure}}}\\
{\footnotesize{Fig. 5: Visual quality comparison of mixed Gaussian-impulse noise removal on image ``Boat". (a) noisy image corrupted by AWGN+SP noise with $\sigma=10$ and $r_{sp}=60\%$. Denoised results by (b) IFASDA [7] (PSNR=27.58 dB, SSIM=0.769); (c) JSM [11] (PSNR=28.06 dB, 0.793); and (d) the proposed De-JASP (PSNR=28.37 dB, SSIM=0.814).}}
\end{figure}

\begin{figure}
{\hspace{1.2cm}
         {\centering
         \begin{subfigure}[b]{0.21\textwidth}
                 \includegraphics[scale=0.35]{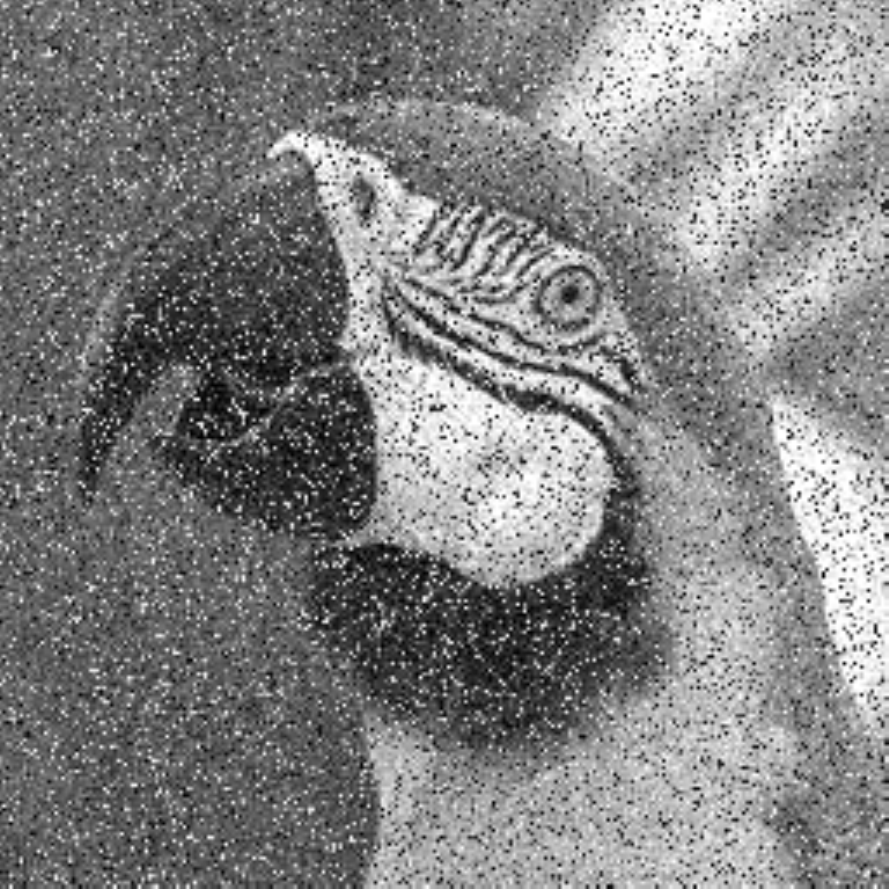}
                  \centering
                 \caption{}
                 \label{fig:1}
         \end{subfigure}%
         \begin{subfigure}[b]{0.201\textwidth}
                 \includegraphics[scale=0.35]{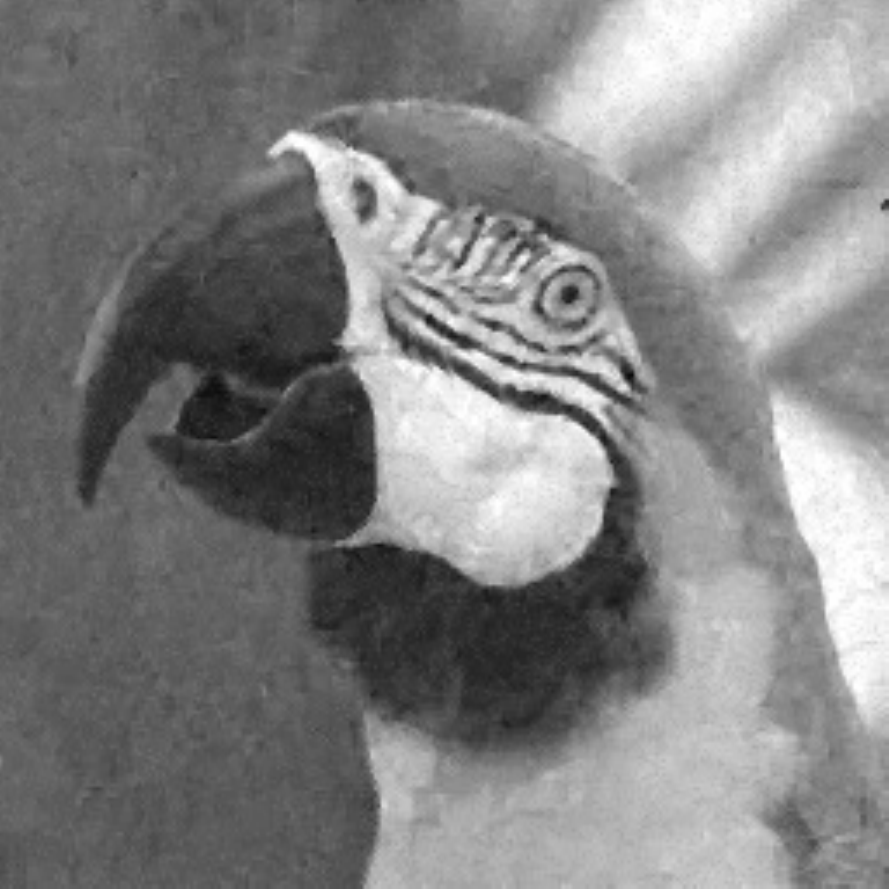}
                 \centering
                 \caption{}
                 \label{fig:1}
         \end{subfigure}
         \begin{subfigure}[b]{0.201\textwidth}
                 \includegraphics[scale=0.35]{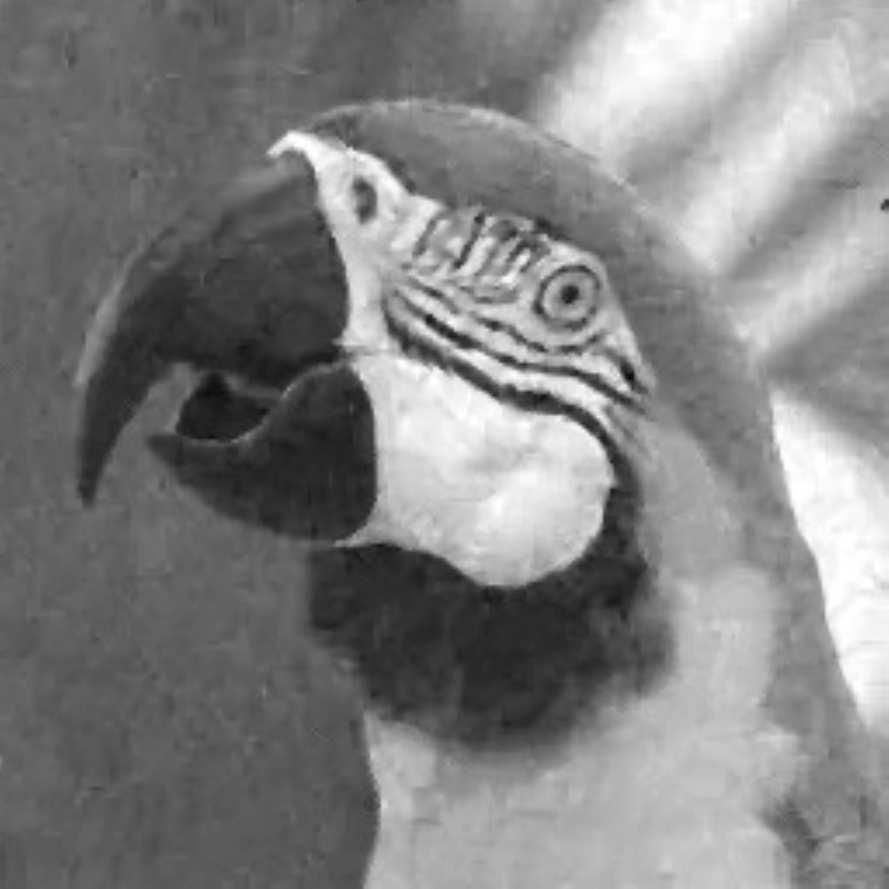}
                  \centering
                 \caption{}
                 \label{fig:1}
         \end{subfigure}%
         \begin{subfigure}[b]{0.21\textwidth}
                 \includegraphics[scale=0.35]{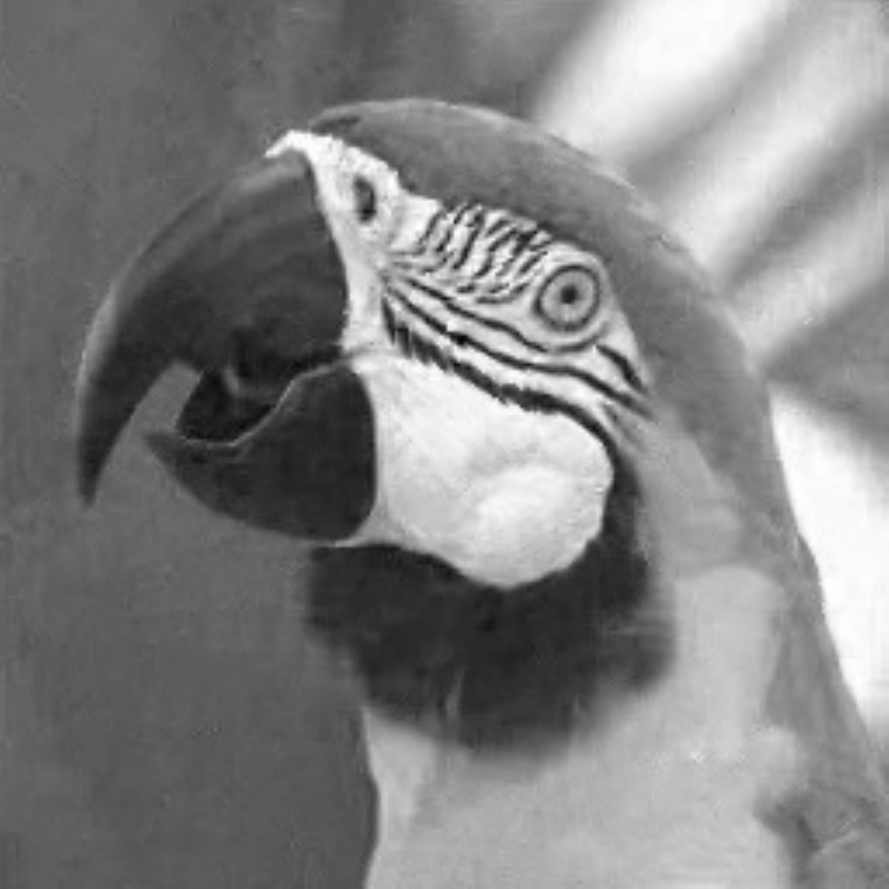}
                 \centering
                 \caption{}
                 \label{fig:1}
         \end{subfigure}}}\\
{\footnotesize{Fig. 6: Visual quality comparison of mixed Gaussian-impulse noise removal on image ``Parrot". (a) noisy image corrupted by AWGN+RV noise with $\sigma=20$ and $r_{rv}=20\%$. Denoised results by (b) IFASDA [7] (PSNR=26.34 dB, SSIM=0.726); (c) JSM [11] (PSNR=26.96 dB, 0.789); and (d) the proposed De-JASP (PSNR=27.42 dB, SSIM=0.816)}}
\end{figure}
\begin{figure}
{\hspace{1.2cm}
         {\centering
         \begin{subfigure}[b]{0.21\textwidth}
                 \includegraphics[scale=0.35]{F_noisy-eps-converted-to.pdf}
                  \centering
                 \caption{}
                 \label{fig:1}
         \end{subfigure}%
         \begin{subfigure}[b]{0.201\textwidth}
                 \includegraphics[scale=0.35]{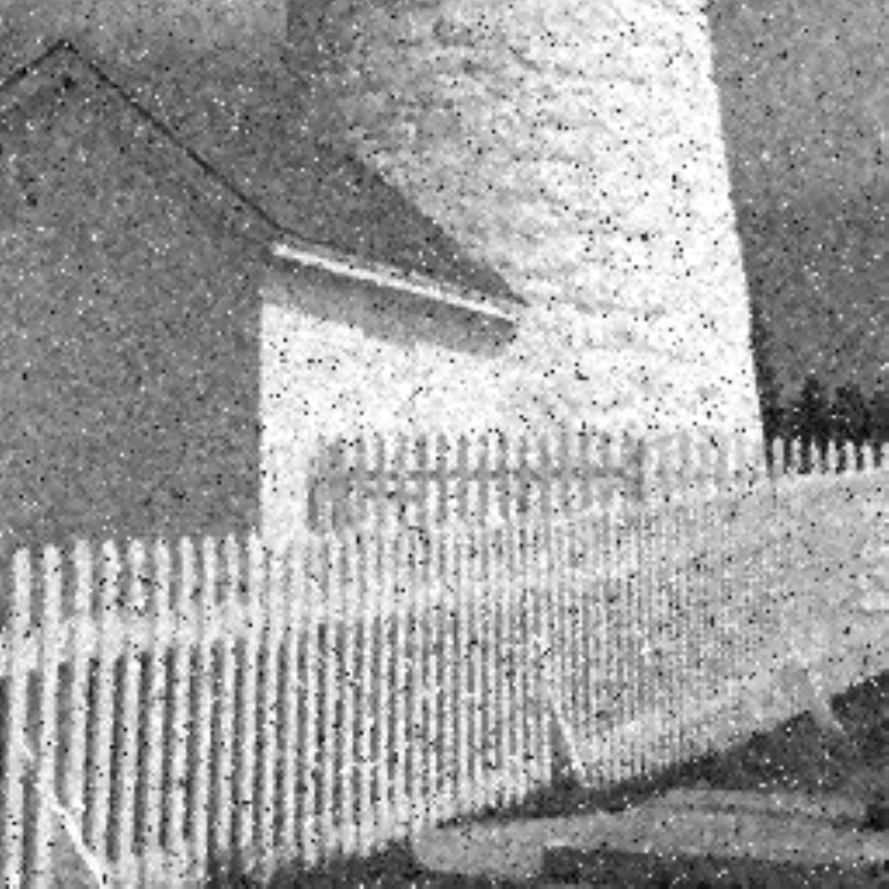}
                 \centering
                 \caption{}
                 \label{fig:1}
         \end{subfigure}
         \begin{subfigure}[b]{0.201\textwidth}
                 \includegraphics[scale=0.35]{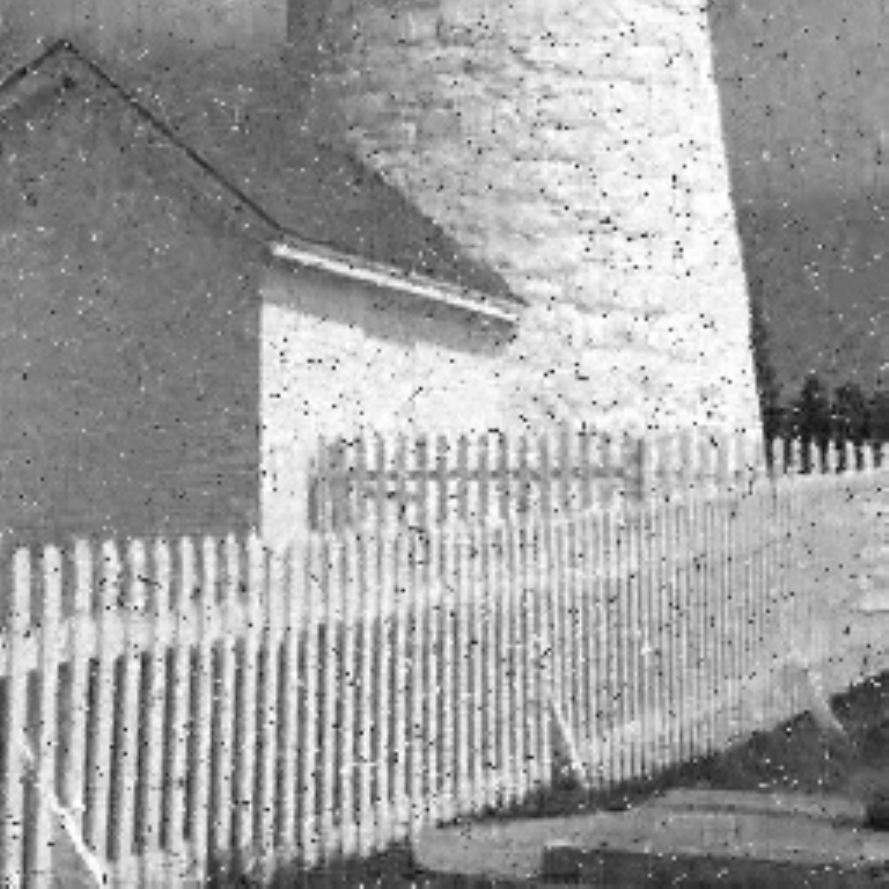}
                  \centering
                 \caption{}
                 \label{fig:1}
         \end{subfigure}%
         \begin{subfigure}[b]{0.21\textwidth}
                 \includegraphics[scale=0.35]{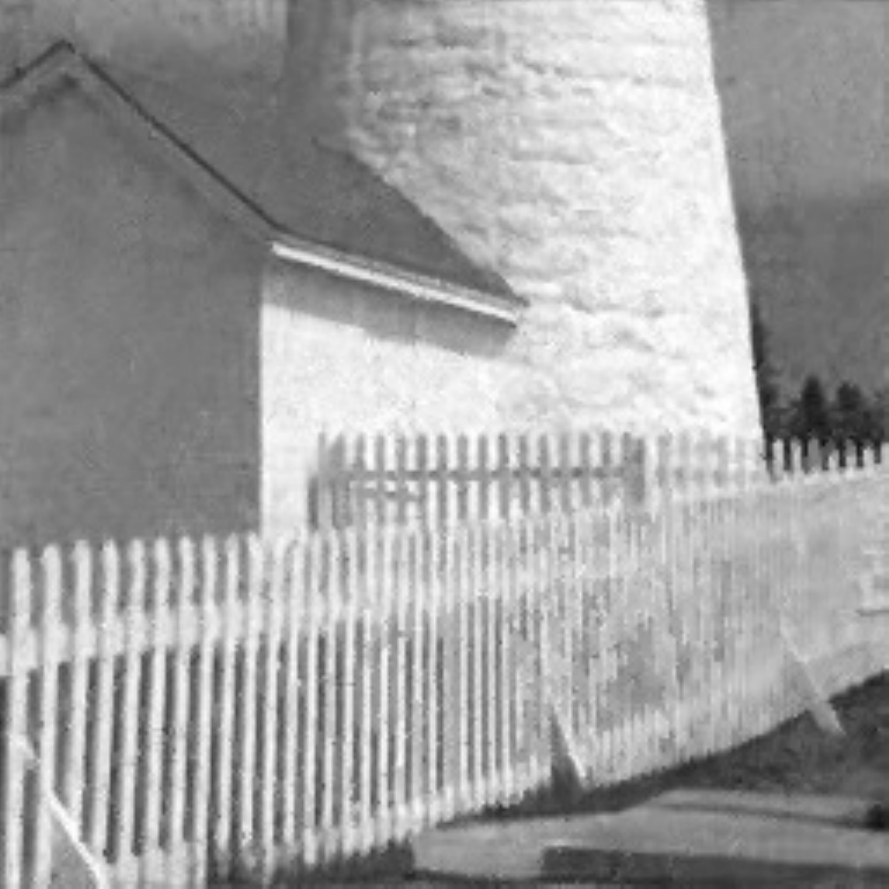}
                 \centering
                 \caption{}
                 \label{fig:1}
         \end{subfigure}}}\\
{\footnotesize{Fig. 7: Visual quality comparison of mixed Gaussian-impulse noise removal on image ``Fence". (a) noisy image corrupted by AWGN+SP+RV noise with $\sigma=10$, $r_{sp}=10\%$ and $r_{rv}=10\%$. Denoised results by (b) IFASDA [7] (PSNR=20.11 dB, SSIM=0.516); (c) JSM [11] (PSNR=21.79 dB, 0.622); and (d) the proposed De-JASP (PSNR=23.19 dB, SSIM=0.698)}}
\end{figure}
\section{Conclusion}
In this paper, first, an adaptive curvelet thresholding (ACT) criterion is introduced to adaptively characterize and abate the perturbations which are appeared during denoising process. Then, a novel joint adaptive statistical prior (JASP) and a new strategy for mixed Gaussian-impulse noise removal via JASP, called De-JASP, is proposed. Extensive experimental results clearly confirm that De-JASP significantly outperforms the current state-of-the-art mixed Gaussian-impulse noise removal techniques.

For future work, we would like to propose an effective nonlocal self-similarity modeling, and pursue this direction to extend our proposed statistical prior to other image restoration applications such as image deblurring and inpainting.

\section*{Acknowledgment}
The authors would like to express their gratitude to Prof. J. Fadili (CNRSENSICAEN-Universit\`{e} de Caen) and Dr. J. Zhang (Peking
University) for many fruitful discussions. They also would like to thank the authors of [7] and [11] for sharing the source code of their papers used in Section V.

\ifCLASSOPTIONcaptionsoff
  \newpage
\fi

\end{document}